\documentclass{llncs}

\usepackage{color}
\usepackage{array,multirow,graphicx}
\usepackage{amsfonts}
\usepackage{tikz}
\usepackage{booktabs}
\usepackage{amsmath}
\usepackage{algpseudocode}
\usepackage{algorithm}

\newcommand{\variables}{\mathcal{V}}
\newcommand{\variable}{v}
\newcommand{\domains}{\mathcal{D}}
\newcommand{\domain}{d}
\newcommand{\constraints}{\mathcal{C}}
\newcommand{\constraint}{c}

\newcommand{\nalog}{\textsc{ESeR}}

\newcommand{\decisions}{\Delta}

\newcommand{\deduction}{\gamma}

\newcommand{\expl}[1]{expl(#1)}

\algnewcommand\algorithmicswitch{\textbf{switch}}
\algnewcommand\algorithmiccase{\textbf{case}}
\algnewcommand\algorithmicassert{\texttt{assert}}
\algnewcommand\Assert[1]{\State \algorithmicassert(#1)}%
\algdef{SE}[SWITCH]{Switch}{EndSwitch}[1]{\algorithmicswitch\ #1\ \algorithmicdo}{\algorithmicend\ \algorithmicswitch}%
\algdef{SE}[CASE]{Case}{EndCase}[1]{\algorithmiccase\ #1}{\algorithmicend\ \algorithmiccase}%
\algtext*{EndSwitch}%
\algtext*{EndCase}%

\newcommand{\IN}{\ensuremath{asg}}
\newcommand{\RM}{\ensuremath{rem}}
\newcommand{\UB}{\ensuremath{upp}}
\newcommand{\LB}{\ensuremath{\ell ow}}
\newcommand{\Evt}{\ensuremath{\Sigma}}
\newcommand{\evt}{\ensuremath{\sigma}}

\newcommand{\dec}{\ensuremath{\delta}}
\newcommand{\Rul}{\ensuremath{\Pi}}
\newcommand{\rul}{\ensuremath{\pi}}

\newcommand{\xp}{\ensuremath{e}}

\newcommand{\choco}{\texttt{choco}}
\newcommand{\cbjfalse}{\texttt{CBJ}}
\newcommand{\cbjtrue}{\texttt{CBJ-i}}

\title{Event Selection Rules to Compute Explanations}
\author{
	Charles~Prud'homme\inst{1}
	\and Xavier~Lorca\inst{1}
	\and Narendra~Jussien\inst{2}
}

\institute{
  TASC (CNRS/INRIA), Mines Nantes, FR -- 44307 Nantes, France \\
  \url{FirstName.LastName@mines-nantes.fr}
    \and
  T\'el\'ecom Lille, FR -- 59653 Villeneuve d'Ascq Cedex, France \\
  \url{Narendra.Jussien@telecom-lille.fr}
}

\begin{document}
\maketitle
\begin{sloppypar}
\begin{abstract}
Explanations have been introduced in the previous century.
Their interest in reducing the search space is no longer questioned.
Yet, their efficient implementation into CSP solver is still a challenge.
In this paper, we introduce \nalog, an Event Selection Rules algorithm that filters events generated during propagation.
This dynamic selection enables  an efficient computation of explanations for intelligent backtracking algorithms.
We show the effectiveness of our approach on the instances of the last three Mini\-Zinc challenges. 
\end{abstract}
\end{sloppypar}
\section{Introduction}
In recent years, Lazy Clause Generation solvers~\cite{DBLP:journals/constraints/OhrimenkoSC09} (through hybrid SAT-CSP) have brought explanations back to attention. Explanations in pure CSP solvers have suffered several drawbacks, despite their unquestioned interest: they usually required a deep refactoring of the targeted solver and computation itself was both memory and CPU consuming. In a few words, they were considered not worth the pain.
In this paper, we claim that explanations have still a bright future in pure CSP solvers. We introduce here a simple yet efficient algorithm for computing explanations for CSP solvers: \nalog~(Event Selection Rules). We exploit event-related information in a novel way in order to focus on those fundamentally related to a failure. 
Combining this technique with asynchronous constraint explanation schemas enables to easily plug intelligent backtracking algorithms in (such as CBJ~\cite{DBLP:journals/ai/Dechter90,prosser95}).
We implemented ESeR in Choco and evaluated it on the instances of the last three Mini\-Zinc challenges. We show its efficiency and practicability.

This paper is organized as follow.
In Section~\ref{bckgnd}, we recall the common background of Constraint Programming. 
The key elements of intelligent backtracking algorithms are introduced in Section~\ref{intback}.
In Section~\ref{eser}, the pseudocode of the \nalog~algorithm is presented.
Comparative evaluations are showed in Section~\ref{eval}.

\section{Background}\label{bckgnd}


A Constraint Satisfaction Problem (CSP) is a triple $\langle \variables, \domains, \constraints \rangle$ where 
$\variables$ is the sequence $(\variable_1,\variable_2, \ldots,\variable_n)$ of \textit{variables},
$\domains$ is the sequence $(\domain_1,\domain_2, \ldots,\domain_n)$ of \textit{domains} associated with the variables,
and $\constraints$ is the set $\{\constraint_1,\constraint_2,\ldots,\constraint_m\}$ of \textit{constraints}.
The domain $\domain_i$ is a finite ordered set of integer values to which the variable $\variable_i$ can be assigned to.
The constraint $\constraint_j$, associates with a sub-sequence $\variables(\constraint_j)\in \variables$, defines the allowed combinations of values which satisfy $\constraint_j$.  
A constraint is equipped with a function $p_{c_j}$ which removes from the domains of $\variables(\constraint_j)$, the values that cannot satisfy $\constraint_j$. 
The goal of a CSP is to find a \emph{solution}. 
A solution is an assignment to all the variables from their respective domains such that all the constraints are simultaneously satisfied.
The exploration of the search space is processed using a \emph{depth-first search}.
At each step, a decision is taken.
Without loss of generality, a (2-way) decision $\dec$ is an assignment of a variable to a value from its domain; its refutation $\neg\dec$ is the removal of this value from the variable domain. 
The satisfiability of all the constraints is then checked thanks to a constraint propagation algorithm which executes $p_{c_j}$ for each constraint in turn until a fixpoint is reached (no more values can be removed).
The propagation can empty a variable domain: a \emph{failure} is detected.
Upon failure, the search backtracks to the last decision and refutes it then resumes search.


How the domain of a variable changes is characterized by \emph{events}~\cite{laburthe.TRICS00,DBLP:journals/toplas/SchulteS08}.
An event $\evt$ is a change in a variable domain such that the cardinality of the resulting domain is strictly smaller.
An event is led by a \emph{cause}: a constraint, a decision or a refutation.
There are four types of events:
a removal ($\RM$) states that a single value is removed from a variable domain,
an assignment ($\IN$) states that all values but one are removed from a variable domain,
a lower bound increase ($\LB$) states that all values below an (exclusive) value are removed from a variable domain, 
and an upper bound decrease ($\UB$) states that all values above an (exclusive) value are removed from a variable domain.
For implementation purpose, additional data is attached to each event depending on its type:
$x$, the value removed from $v$, for $\RM$,
$\ell_o$ and $\ell_n$, respectively the previous and the new lower bound of $v$, for $\LB$,
$u_o$ and $u_n$, respectively the previous and the new upper bound of $v$, for $\UB$,
and $a$, $\ell_o$ and $u_n$, respectively the instantiated value of $v$, its previous lower and upper bound, for $\IN$.

\section{Intelligent backtracking}\label{intback}
Nogoods and explanations have been used for a long time to improve search~\cite{DBLP:journals/jair/Ginsberg93,DBLP:journals/endm/JussienL00,Katsirelos09,prosser95,DBLP:journals/constraints/PrudhommeLJ14}.
They enable doing intelligent backtracking (non-chronological) and can reduce significantly the number of decisions taken and speed up the resolution. 
Moreover, detecting which partial assignments cannot be extended to a solution avoids reproducing them in the remainder of the search tree.
This is related to the notion of \emph{nogood}.
\begin{definition}[Nogood~\cite{DBLP:journals/ai/Dechter90}\label{labelng}]
A \emph{nogood} is a set a of assignments that cannot all be true in any solution of the CSP.
\end{definition}
The definition~\ref{labelng} can be generalized to contain either assignments or removals.
\begin{definition}[Generalized Nogood~\cite{Katsirelos09}]
A \emph{g-nogood} is a set a of assignments and removals that cannot all be true in any solution of the CSP.
\end{definition}
Katsirelos~\cite{Katsirelos09} designed \emph{$1^{st}$-Decision scheme}, an algorithm to learn g-nogoods during the search. 
First, each event caused by a constraint is labeled with a g-nogood  and the depth at which it was made, until a failure occurred.
The g-nogood is the result of a direct constraint violation. 
Decisions are labeled with \emph{null}.  
Then, exploring events in a bottom-up way iteratively modifies the g-nogood derived from the failure:
the deepest event is replaced by its g-nogood, until the deepest event to replace is a decision.
The resulting g-nogood is used to backtrack and to label the refutation upon backtrack. 
Finally, it is posted into the solver to avoid exploring the same search subtree in the future.  
Gent et al.~\cite{DBLP:conf/padl/GentMM10} suggest not to store the g-nogood
but a polymorphic function able to retrieve it from a given event. 
The g-nogood can thus be calculated lazily. They also show how such a function can be defined for various constraints. 
It turned out to be very effective in practice by avoiding the calculation of useless g-nogoods and thus, by saving time. 
Note that 
$1^{st}$-Decision scheme does not exploit events on bounds and processes them as a sequence of removals. 

A trivial nogood can be obtained from the set of decisions \emph{explaining} the failure.
This is the role of \emph{explanations} whose definition is related to \emph{deductions}.
\begin{definition}[Deduction~\cite{jussien:tel-00293905}]
A \emph{deduction} $\deduction_{\evt}$ is the determination that an event $\evt$ should be applied,
for instance, that a value should be removed the domain of a variable. 
\end{definition}


\begin{definition}[Explanation~\cite{jussien:tel-00293905}]
An \emph{explanation} $\expl{\deduction_\evt}$ for the deduction $\deduction_\evt$ is defined by 
a set of constraints $\constraints'\subseteq\constraints$ 
and a set of decisions $\decisions$, 
such that $\constraints'\land\decisions\land\neg\evt$ cannot all be true in a solution.
\end{definition}

Jussien and Barichard~\cite{Jussien00thepalm} introduced PaLM, an explanation based constraint programming system.
In PaLM, an explanation is computed and attached when an event is generated.
The code of the constraints are modified in consequence, but low-intrusive techniques can be implemented, such as the ones described in~\cite{jussien:tel-00293905}.
The computation may be done as and when the events are generated or by storing a list of event-related information and compiling them into explanations on demand. 
Then, the explanation of a failure is the union of the explanations of each value removed from the empty domain.
Explanation is the key point of intelligent backtracking algorithms, such as Conflict-directed Backjumping~\cite{DBLP:journals/ai/Dechter90,prosser95}, Dynamic Backtracking~\cite{DBLP:journals/jair/Ginsberg93,jussien-macdbt-cp}, Path Repair~\cite{DBLP:journals/endm/JussienL00,DBLP:conf/cpaior/PraletV04} or Explanation-based LNS~\cite{DBLP:journals/constraints/PrudhommeLJ14}. However, the memory footprint and the CPU consumption turn out to be the major concerns of using explanations in a CSP solver.
In the following, we describe an algorithm that improves the computation of explanations in a CSP solver.

\section{Event selection rules to compute explanations}\label{eser}
We now present \textsc{E}vent \textsc{Se}lection \textsc{R}ules (\nalog{}), an algorithm based on the ability to determine relevant event-based information.
It exploits the types of modification variables can have and relies on an asynchronous computation of explanations.  
We first present the pre-requisites, and then describe the notion of \emph{event selection rule}.
Finally, we describe the algorithm itself.

\subsubsection{Pre-requisites}
To be able to evaluate the events, we have to store them in a chronologically ordered list, named $\Evt$, during the propagation.
Keeping the events sorted is necessary to recognize the inheritance between them.
The index of the last event stored is backtrackable, thus, outdated events are automatically forgotten on backtrack.
The cause which produced an event $\evt$ has an \emph{explanation schema}, or e\nobreakdash-schema, which is able to point out which (earlier) events may be the source of $\evt$.
A decision is event-independent and thus points out no event,  
a refutation is explained by previously taken decisions~\cite{jussien:tel-00293905}.  


\subsection{Event selection rules}
When a domain becomes empty, the question the explanation has to answer is \emph{what are the causes and events related to those value removals ?}
In other words, how to recognize relevant events which have led, by propagation, to remove all the values from a specific domain ? 
To do so, we introduce \emph{event selection rules}.
An event selection rule $\rul$ characterizes that a certain modification which occurred on variable $v$ is relevant to explain the failure. 
It can be seen as a \emph{negative} of an event.
The type of modifications can be:
\begin{description}
	\item[$\rul_r$:] the event which removed a specific value from $v$ is relevant,
	\item[$\rul_u$:] the events which modified the lower bound of $v$ are relevant,
	\item[$\rul_\ell$:] the events which modified the upper bound of $v$ are relevant,
	\item[$\rul_d$:] any events occurring on $v$ are relevant.
\end{description}
In the event list $\Evt$, the $\LB$ events (resp. $\UB$ events) are naturally ordered in such way that the lower of bound of a given variable only increases (resp. decreases).
This is not true for $\RM$ events, that is why an integer is attached to a $\rul_r$ rule representing the removed value of $v$, denoted $\rul_r.x$ in the following.  
If an event is relevant, the e\nobreakdash-schema of the cause is interrogated.
A e\nobreakdash-schema can declare new event selection rules which will point out to earlier events.   
For instance, when the domain of the variable $v_1$ becomes empty, the first rule declared is $\langle v_1, \rul_d \rangle$ which will select all events involving $v_1$.

The \nalog{} function, described in Algorithm~\ref{alg:1}, is called when a variable domain becomes empty.
The affected variable $v$, the list of events $\Evt$ and a boolean $pe$ are passed as parameters of the function. 
By setting $pe$ to $false$, all events from $\Evt$ are visited and a classical explanation is computed.
By setting that parameter to $true$, though, the analysis of $\Evt$ stops as soon as a decision satisfies a rule and an incomplete explanation is returned.
Computing classical explanations is only required for some intelligent backtracking algorithms (discussed in Section~\ref{iback}). 
First, an empty explanation $\xp$ is created (line~2).
The set of event selection rules is initialized with the rule derived from the empty domain variable (line~3).
Then, a bottom-up analysis of events in $\Evt$ is run in order to find those satisfying a rule of $\Rul$ (lines~5-23).
If an event satisfies a rule from $\Rul$ (line~8), then the explanation of the failure needs to be updated as well as, possibly, the set of rules.
These operations are executed while all the events from $\Evt$ have not been visited or a stop condition is encountered (line~6).
When the iteration ends and if incomplete explanations are allowed, the rules gathered up to this point are copied into the explanation (line~24-25).
In this way, the computation may be resumed, if necessary (see Section~\ref{iback}).
Note that, merging two explanations (line~13) also merges rules, if any.
The computed explanation is then returned (line~27).

\begin{algorithm}[]
\begin{algorithmic}[1]
\Function{explain}{an empty domain variable: $v$, a list of events: $\Evt$, boolean pe}
\State $\xp \leftarrow \emptyset$ \Comment{Create a new explanation} 
\State $\Rul \leftarrow \langle v, \rul_d \rangle$ \Comment{Initialize the rules set}
\State $stop \leftarrow false$
\State $i = |\Evt|-1$
\While{$i \geq 0$ and $\neg stop$}
\State $\evt \leftarrow \Evt[i]$
\If{$\evt$ satisfies a rule of $\Rul$} \Comment{See Table~\ref{tab:cov}}
\If{$\evt.c$ is a decision}
\State add the decision to $\xp$
\State  $stop \leftarrow pe$
\ElsIf{$\evt.c$ is a refutation}
\State merge the explanation of the refutation into $\xp$
\Else \Comment{The cause is a constraint}
\State add the constraint to $\xp$
\State update $\Rul$ with rules returned by the constraint's e\nobreakdash-schema 
\EndIf
\If{$\evt.t = \RM$}
\State remove the corresponding rule from $\Rul$
\EndIf
\EndIf
\State $i = i -1$ 
\EndWhile
\If{pe} \State copy rules in $\xp$
\EndIf
\State \Return $\xp$
\EndFunction
\end{algorithmic}
\caption{\nalog{}: Event Selection Rule}\label{alg:1}
\end{algorithm}
\begin{table}[]
\caption{Covering conditions to select an event.}
\begin{center}
\begin{tabular}{cccccc}
\toprule
 & & \multicolumn{4}{c}{\textbf{event's type}}\\
 & &\multicolumn{1}{c}{$\IN$} & \multicolumn{1}{c}{$\LB$} & \multicolumn{1}{c}{$\UB$} & \multicolumn{1}{c}{$\RM$}\\
\midrule
 \parbox[t]{4mm}{\multirow{5}{*}{\rotatebox[origin=c]{90}{\textbf{rule's type}}}}
 & $\rul_d$ &$true$&$true$&$true$&$true$\\
 & $\rul_\ell\land\rul_u$ & $true$&$true$&$true$&$\evt.r \not \in d_{\evt.v}$\\
 & $\rul_\ell$ &
 $\evt.r \not \in d_{\evt.v}$
 &$true$&$false$&$\evt.r \not \in d_{\evt.v}$\\
 & $\rul_u$ &
 $\evt.r \not \in d_{\evt.v}$
 &$false$&$true$&$\evt.r \not \in d_{\evt.v}$\\
 & $\rul_r$ &$\rul_r.x \in [\![\evt.\ell_o,\evt.u_o]\!]$&$\rul_r.x \in [\![\evt.u_n,\evt.u_o]\!]$&$\rul_r.x \in [\![\evt.\ell_o,\evt.\ell_n]\!]$&$\rul_r.x  = \evt.r$\\
\bottomrule
 \end{tabular}
\end{center}
\label{tab:cov}
\end{table}%

The satisfaction test (Algorithm~\ref{alg:1}, line~8) simply verifies if  
a rule based on the event's variable exists
and if the event's modification type is covered by the rule's type, which may imply an evaluation of $\evt.X$ and $\rul_r.x$.
The covering conditions are depicted in Table~\ref{tab:cov}.
For instance, a $\rul_d$ selection rule exists for a variable $v_1$ covers a $\LB$ event for $v_1$.

When updating the explanation (Algorithm~\ref{alg:1}, line~9-21), three cases are considered:
 (a) the cause is a decision (line~9-11), (b) the cause is a refutation (line~12-13) and (c) the cause is a constraint (line~15-16).
In (a), the decision is added to $\xp$. Then, if an incomplete explanation is allowed, $stop$ is set to $true$ in order to interrupt the iteration over events.
In (b), a previous failure led to refute a decision whom explanation was previously computed.
That explanation is then merged into the current one. 
In (c) the constraint is added to $\xp$.
Then, the e\nobreakdash-schema of the constraint is interrogated to add new rules to $\Rul$.
A valid e\nobreakdash-schema for all constraint is to declare a $\rul_d$ rule for all variables of the constraint.  \label{pseudoc}
However, it results in weak explanation, and specific e\nobreakdash-schemas should be implemented for each constraint.
Finally, the set of rules $\Rul$ can be reduced whenever a $\rul_r$ rule is satisfied, since it can appear only once in a search branch (line~18-20). 

The \nalog{} algorithm runs in $\mathcal{O}(|\Evt|\times s)$ where $s$ is complexity of the most complex e\nobreakdash-schema, at each failure.
Indeed, the while-loop (line~8-23) only executes $\mathcal{O}(1)$ operations except when it interrogates the explanation schema of the cause (line~16).
The complexity of determining which rules need to be added to the rules set depends on the type of cause.

\subsection{Using \nalog{} in intelligent backtracking algorithms}\label{iback}
Intelligent backtracking algorithms can be split in three groups.
First, those requiring classical explanations, such as Path Repair or Explanation-based LNS. 
They rely on a relaxation of the explaining decisions set, so all of them must be retrieved to be correct.
Second, those accepting incomplete explanations such as Conflict-directed Backjumping.
Only the deepest decision from the explaining ones is interesting to jump back to it, so finding it is enough. 
Finally, those accepting incomplete explanations with the necessity to keep up their computation later, such as Dynamic Backtracking.
Dynamic Backtracking mimics Conflict-directed Backjumping, but when jumping the deepest decision, decisions and refutations taken between the failure and the one to jump back to have to be kept by the search heuristic. 
Refutations depend on earlier decisions and, potentially, on the one to jump back to.
Stating whether or not a refutation depends on the jump back decision is made by keeping up their computation until the event corresponding to that decision is visited.
If the decision satisfies one of the rules, the refutation depends on the jump back decision.

\paragraph{Implementation}
For implementation concerns, all rules involving the same variable are represented with a unique object made of two integers and an integers set.
The two integers correspond to the variable index and the aggregated modification types (handled with bitwise operations).
The integers set manages removed values.
An explanation is composed of a decisions set (encoded in a bitset), a constraints set and a rules set, possibly empty.
Embedding a rules set in the explanation is motivated to keep up computing it, which is required when dealing with incomplete explanations.
Finally, the rules maintained in the \nalog{} algorithm can be compiled in order to extract g-nogoods.
For efficiency purpose, the g-nogood must be reviewed to deal with bound modifications, in such way that they are not handled as a sequence of value removals but as native cases.
Extracting g-nogoods from the rules set is not implemented in the evaluated version of \nalog.

\section{Evaluation}\label{eval}

Our explanation engine has been implemented in Choco-3.3.1\footnote{Available on \url{www.choco-solver.org}}~\cite{choco}, a Java library for constraint programming.
We evaluated three configurations: 
a standard backtrack search (\choco), 
a conflict-directed backjumping search disabling incomplete explanations (\cbjfalse) and CBJ enabling incomplete explanations (\cbjtrue).
Not all (global) constraints are natively explained in Choco.
If not, the naive but valid  e\nobreakdash-schema depicted in Section~\ref{pseudoc} is used.
The set of benchmarks used consists of the instances from the Mini\-Zinc Challenges\footnote{\url{http://www.minizinc.org/challenge.html}} (2012, 2013 and 2014), and it is composed of 269 instances from 66 problems.
The models rely on a large variety of constraints, including global constraints, and each problem describes a dedicated search strategy.
Each of the 269 instances was executed with a 15-minute timeout, on a Macbook Pro with 6-core Intel Xeon at 2.93Ghz running a MacOS 10.10, and Java 1.8.0\_05.
Each instance was run on its own core, each with up to 4096MB of memory.
Note that because there are too many distinct types of problems (66), we did not characterize them on the scatterplots.

\begin{figure}[htpd]
\centering
\includegraphics[scale=.5,trim = 60 130 0 150]{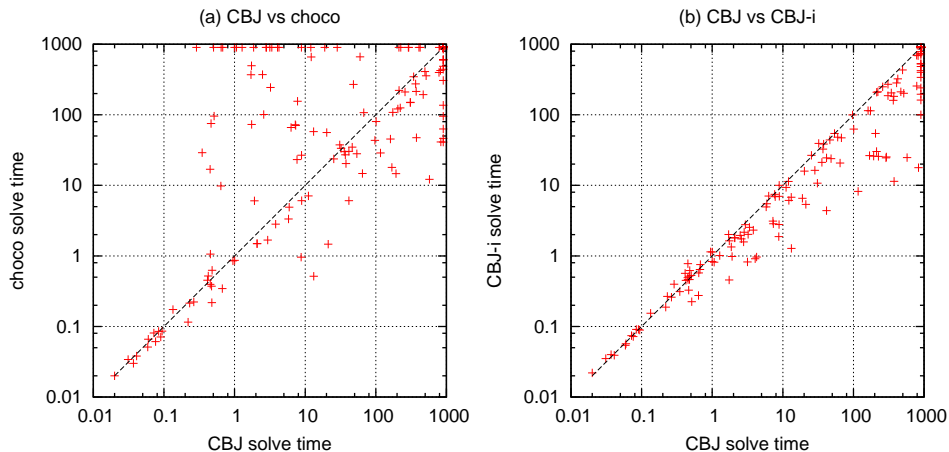}
\caption{Comparative runtime evaluations. 
Each point corresponds to a single instance.
The $x$-axis is the solve time (in seconds) for a first approach, the $y$-axis is the solve time (in seconds) for the second one.
A point above the diagonal means the first approach was faster, below the diagonal that the second one was faster. 
\label{fig1}
}
\end{figure}

Figure~\ref{fig1} reports the comparative evaluations: (a) \cbjfalse{} vs. \choco{} on the left plot and (b) \cbjfalse{} vs. \cbjtrue{} on the right plot.
Note that the axes are in logarithmic scale.
On plot (a), the contribution of using explanations on some certain instances is clear, it corresponds to the points above the diagonal.
Even if almost 50\% of instances timed out for both approaches, \cbjfalse{} is faster in 46\% of the remaining instances, and the speedups can go up to  $286$x.
On plot (b), the benefit of using incomplete explanations is without any doubts in favor of \cbjtrue{}: most of the points are below the diagonal.
Here again, up to 50\% of the instances timed out for the two approaches but enabling incomplete explanation is particularly remarkable:  79\% of the remaining instances are solved faster with \cbjtrue{}. 
The speedups can go up to $47$x.
We do not report the comparative evaluations of \choco{} and Dynamic Backtracking, but the results (available on request) are very close to those observed with CBJ.

%

\section{Conclusion}\label{conclu}
In this paper, we introduced \nalog, an Event Selection Rules algorithm for CSP solvers that filters events generated during propagation.
This dynamic selection enables an efficient computation of explanations for intelligent backtracking algorithms.
The efficiency of our approach has been validated, in practice, on a dataset of 269 instances from the three last MiniZinc challenges.
The further works include an improvement of the possibilities offered by \nalog, by extracting g-nogoods from the selection rules, and a study of alternative intelligent backtracking algorithms.
\bibliographystyle{plain}
\bibliography{eff_expl}

\end{document}